# Comparison and Adaptation of Automatic Evaluation Metrics for Quality Assessment of Re-Speaking

Krzysztof Wołk, Danijel Koržinek
Department of Multimedia
Polish - Japanese Academy of Information Technology.
`kwolk@pja.edu.pl, danijel@pja.edu.pl`

**Abstract**

Re-speaking is a mechanism for obtaining high quality subtitles for use in live broadcast and other public events. Because it relies on humans performing the actual re-speaking, the task of estimating the quality of the results is non-trivial. Most organisations rely on humans to perform the actual quality assessment, but purely automatic methods have been developed for other similar problems, like Machine Translation. This paper will try to compare several of these methods: BLEU, EBLEU, NIST, METEOR, METEOR-PL, TER and RIBES. These will then be matched to the human-derived NER metric, commonly used in re-speaking.

## 1 Introduction

One of the main driving forces in Speech Technology, for the last several years, comes from the efforts of various groups and organizations tackling with the issue of disability, specifically deaf and hard of hearing people. Most notably, a long term effort by such organisations has lead to a plan by the European Commision to enable "Subtitling of 100% of programs in public TV all over the EU by 2020 with simple technical standards and consumer friendly rules" [15]. This ambitious task would not be possible to achieve without the aid of Speech Technology.

While there has been a considerable improvement of quality of Automatic Speech Recognition (ASR) technology recently, many of the tasks present in real-life are simply beyond complete automation. On the other hand, there are tasks, which are also impossible to achieve by humans without the aid of ASR. For example, movie subtitles are usually done by human transcribers and can take a day, up to a week, per material to complete. Live subtitling, however, can sustain only a few seconds delay between the time an event is recorded and the time it appears on the viewer's screen. This is where re-speaking comes into play.

The idea of re-speaking is to use ASR to create live and fully annotated subtitles, but rather than risking misrecognition of the speech happening in the live recording, a specially trained individual, the so-called re-speaker, repeats the speech from the recorded event in a quiet and controlled environment. This approach has many advantages that guarantee excellent results: controlled environment, the ability to adapt the speaker, the ability of the speaker to adapt to the software, solving problems like double-speak, cocktail party effect and non-grammatic speech by paraphrasing. The standard of quality required by many jurisdictions demands less than 2% of errors [17]. From the point of view of a re-speaker, this problem is very similar to that of an interpreter only instead of translating from one language to another, re-speaking is usually done within one language only. There are many aspects of re-speaking worthy of their own insight [16], but this paper will deal only with the issue of quality assessment.

Similarly to Machine Translation (MT), the assessment of the accuracy of re-speaking is not a trivial task, because there are many possible ways to paraphrase an utterance, just like there are many ways to translate any given sentence. Measuring the accuracy of such data has to take semantic meaning into account, rather than blindly performing simple word-to-word comparison.

One option is to use humans to perform this evaluation, as in the NER model, described later in the paper. This has been recognized as very expensive and time-consuming [8]. As a result, human effort cannot keep up with the growing and continual need for the evaluation. This led to the recognition that the development of automated evaluation techniques is critical. [8,9]

Unfortunately most of the automatic evaluation metrics were developed for other purposes than re-speaking (mostly machine translation) and are not suited for languages like Polish that differ semantically and structurally from English. Polish has complex declension, 7 cases, 15 gender forms, and complicated grammatical construction. This leads to a larger vocabulary and greater complexity in data requirements for such tasks. Unlike Polish, English does not have declensions. In addition, word order, esp. the Subject-Verb-Object (SVO) pattern, is absolutely crucial to determining the meaning of an English sentence [10]. While these automatic metrics have already been thoroughly studied, we feel that there is still much to be learnt, especially in different languages and for different tasks, like re-speaking.

This paper will compare some of these automated and human assisted metrics, while also considering issues related specifically to Polish. A small introduction to all the metrics is presented in the beginning of the paper, followed by an experiment using actual re-speaking data.

## 2 Machine Translation Metrics

### 2.1 BLEU Metric

BLEU was developed based on a premise similar to that used for speech recognition, described in [1] as: "The closer a machine translation is to a professional human translation, the better it is." So, the BLEU metric is designed to measure how close SMT output is to that of human reference translations. It is important to note that translations, SMT or human, may differ significantly in word usage, word order, and phrase length. [1]

To address these complexities, BLEU attempts to match variable length phrases between SMT output and reference translations. Weighted match averages are used to determine the translation score. [2]

A number of variations of the BLEU metric exist. However, the basic metric requires calculation of a brevity penalty PB, which is calculated as follows:

$$P_B = \begin{cases} 1, c > r \\ e(1^{-rc}), c \leq r \end{cases}$$

where r is the length of the reference corpus, and candidate (reference) translation length is given by c. [2]

The basic BLEU metric is then determined as shown in [2]:

$$BLEU = P_B \exp\left(\sum_{n=0}^{N} w_n \log p_n\right)$$

where $w_n$ are positive weights summing to one, and the n-gram precision $p_n$ is calculated using n-grams with a maximum length of N.

There are several other important features of BLEU. First, word and phrase position within the text are not evaluated by this metric. To prevent SMT systems from artificially inflating their scores by overuse of words known with high confidence, each candidate word is constrained by the word count of the corresponding reference translation. A geometric mean of individual

sentence scores, with consideration of the brevity penalty, is then calculated for the entire corpus. [2]

## 2.2 NIST Metric

The NIST metric was designed to improve BLEU by rewarding the translation of infrequently used words. This was intended to further prevent inflation of SMT evaluation scores by focusing on common words and high confidence translations. As a result, the NIST metric uses heavier weights for rarer words. The final NIST score is calculated using the arithmetic mean of the n-gram matches between SMT and reference translations. In addition, a smaller brevity penalty is used for smaller variations in phrase lengths. The reliability and quality of the NIST metric has been shown to be superior to the BLEU metric in many cases. [3]

## 2.3 Translation Edit Rate (TER)

Translation Edit Rate (TER) was designed to provide a very intuitive SMT evaluation metric, requiring less data than other techniques while avoiding the labor intensity of human evaluation. It calculates the number of edits required to make a machine translation match exactly to the closest reference translation in fluency and semantics. [4, 5]

Calculation of the TER metric is defined in [4]:

$$TER = \frac{E_w}{R}$$

where E represents the minimum number of edits required for an exact match, and the average length of the reference text is given by wR. Edits may include the deletion of words, word insertion, word substitutions, as well as changes in word or phrase order. [4]

## 2.4 METEOR Metric

The Metric for Evaluation of Translation with Explicit Ordering (METEOR) is intended to take several factors that are indirect in BLEU into account more directly. Recall (the proportion of matched n-grams to total reference n-grams) is used directly in this metric. In addition, METEOR explicitly measures higher order n-grams, considers word-to-word matches, and applies arithmetic averaging for a final score. Best matches against multiple reference translations can also be used. [5]

The METEOR method uses a sophisticated and incremental word alignment method that starts by considering exact word-to-word matches, word stem matches, and synonym matches. Alternative word order similarities are then evaluated based on those matches.

Calculation of precision is similar in the METEOR and NIST metrics. Recall is calculated at the word level. To combine the precision and recall scores, METEOR uses a harmonic mean. METEOR rewards longer n-gram matches. [5]

The METEOR metric is calculated as shown in [5]:

$$\text{METEOR} = \left(\frac{10\,P\,R}{R + 9\,P}\right)(1 - P_M)$$

where the unigram recall and precision are given by R and P, respectively. The brevity penalty $P_M$ is determined by:

$$P_M = 0.5\left(\frac{C}{M_U}\right)$$

where $M_U$ is the number of matching unigrams, and C is the minimum number of phrases required to match unigrams in the SMT output with those found in the reference translations.

An important factor in METEOR is the liberal use of weights for the numerous aspects of the system. These weights need to be tuned for specific tasks to match human judgment precisely. The tool comes with a couple of pre-tuned parameter sets for some common task and language pairs, but these obviously don't include either re-speaking or Polish.

## 2.5 METEOR-PL

While often used in a language-independent manner, the greatest advantage of METEOR is its ability to model features of a specific language, like the aforementioned synonyms, stems and paraphrasing. These features are enabled through the use of standard language tools, easily obtainable for many languages. In order to adapt METEOR to Polish, several steps needed to be made.

The synonym matcher uses a special script for extracting relevant information from the Princeton WordNet [11] project. In Polish, there is an equivalent project developed at the Wrocław University of Technology [12] and it works exactly the same as the Princeton original.

The standard METEOR stemmer is implemented using the Snowball [13] tool, but this only supports a limited set of languages. Other languages can be implemented by hand-crafting rules using a special finite grammar. For Polish, a well known morphological analyzer and stemmer Morfologik [14] developed at the IPI PAN was used instead. This meant that the METEOR source code needed to be slightly modified, specifically to support multiple stems per word.

One final modification of the system, with respect to Polish, was the creation of a list of function words. This list isn't very big (this is true for other languages as well) and may have to be adjusted for specific uses. At its current state it contains punctuation, some common abbreviations and common conjunctions.

The last feature of METEOR is its ability to model paraphrasing. This uses a system that is trained on a parallel corpus. At the moment of writing this paper, the amount of respeaking data was so small that no such corpus could be readily produced. This is something that could significantly improve the measure in the future, however.

## 2.6 RIBES

The focus of the RIBES metric is word order. It uses rank correlation coefficients based on word order to compare SMT and reference translations. The primary rank correlation coefficients used are Spearman's $\rho$, which measures the distance of differences in rank, and Kendall's $\tau$, which measures the direction of differences in rank. [6]

These rank measures can be normalized to ensure positive values [6]:
Normalized Spearman's $\rho$ $(NSR) = (\rho + 1)/2$
Normalized Kendall's $\tau$ $(NKT) = (\tau + 1)/2$

These measures can be combined with precision P and modified to avoid overestimating the correlation of only corresponding words in the SMT and reference translations:

$$NSR\ P\alpha\ and\ NKT\ P\alpha$$

where α is a parameter in the range 0 < α < 1.

## 2.7 EBLEU

We now discuss enhancements to the BLEU metric that we introduced in [7]. The goal was to make this metric more reliable when it comes to morphologically rich languages like Polish. In particular, our enhanced metric rewards synonyms and rare word translations, while modifying the calculation of cumulative scores.

### a. Consideration of Synonyms

In our enhanced metric, we would like to reward matches of synonyms, since the correct meaning is still conveyed.

Consider this test phrase: "this is a exam" and this reference phrase: "this is a quiz"

The BLEU score is calculated as follows:

$$BLEU = (1+1+1+0)/4 = 3/4 = 0.75$$

BLEU does not count the word "exam" as a match, because it does not find it in the reference phrase. However, this word is not a bad choice. In our method, we want to score the synonym "exam" higher than zero and lower than the exact word "quiz".

To do this, for each word in a test phrase we try to find its synonyms. We check for an exact word match and for all test phrase synonyms to find the closest words to the reference.

We apply the default BLEU algorithm to the modified test phrase and reference phrase, with one difference. The default BLEU algorithm scores this new test phrase as 1.0, but we know that the original test phrase is "this is a exam". So, we would like to give a score higher than 0.75 but less than 1.0 to the test phrase.

During the BLEU evaluation, we check each word for an exact match. If the word is a synonym and not an exact match, we do not give a full score to that word. The score for a synonym will be the default BLEU score for an original word multiplied by a constant (synonym-score).

For example, if this constant equals 0.90, the new score with synonyms is:

$$(1+1+1+0.9)/4 = 3.9/4 = 0.975$$

With this algorithm, we have synonym scores for all n-grams, because in 2-gram we have "a quiz" and in 3-gram, "is a quiz" in both test and reference phrases.

### b. Consideration of Rare Words

Our algorithm gives extra points to rare word matches. First, it obtains the rare words found in the reference corpus. If we sort all distinct words of the reference with their repetition order (descending), the last words in this list are rare words. The algorithm takes a specific percentage of the whole sorted list as the rare words (rare-words-percent).

When the default BLEU algorithm tries to score a word, if this word is in the rare word list, the score is multiplied by a constant (rare-words-score). This action applies to all n-grams. So, if we have a rare word in a 2-gram, the algorithm increases the score for this 2-gram. For example, if the word "roman" is rare, the "roman empire" 2-gram gets an increased score. The algorithm is careful that score of each sentence falls within the range of 0.0 and 1.0.

### c. Determination of Cumulative Score

The cumulative score of our algorithm combines default BLEU scores using logarithms and exponentials as follows:
1. Initialize s = 0
2. For each $i$th-gram:
a.   s = s + log($B_i$)
b.   $C_i$ = exp(s / $i$)

where $B_i$ is the default BLEU score and $C_i$ is the cumulative score.

In addition, we know that:

$$\exp(\log(a) + \log(b)) = a * b$$

and:

$$\exp(\log(a)/b) = a^{(\frac{1}{b})}$$

# 3 NER Subtitle Accuracy Model

The NER model [17] is a simple extension of the word accuracy metric adapted specifically for measuring the accuracy of subtitles. It is one of two measures that is of particular importance for media providers (television companies, movie distributors, etc), the other one being the reduction rate. Generally, the aim of good subtitles is to reduce the length of written text as much as possible (in order to preserve space on screen and make it easier to read) while maintaining an almost perfect accuracy (usually above 98%).

Since we are dealing with paraphrasing, it is very difficult to perform accurate measurements by comparing the text only. The NER model gets around this problem by counting errors using a simple formula, which inspired its name:

$$\text{NER accuracy} = \frac{N - E - R}{N} \times 100\%$$

Where N is the number of analyzed tokens (usually also includes punctuation), E is the number of errors performed by the re-speaker, and R is the number of errors performed by the ASR system (on re-speaker's speech). Additionally, the errors in E are weighted: 0.25 for minor errors, 0.5 for normal and 1 for serious errors. There are user-friendly tools available for making these assessments and obviously there may be a certain level of human bias involved in these estimates. Nevertheless, all the decisions are thoroughly checked and explainable using this method, which makes it one of the most popular techniques for subtitle quality assessment used by many regulatory bodies worldwide.

# 4 Dataset and Experiment Setup

The data used in the experiments described in this paper was collected during a study performed in the Institute of Applied Linguistics at the University of Warsaw [16]. This, still ongoing, study aims to determine the relevant features of good re-speakers and their technique. One of the obvious measures is naturally the quality of re-speaking discussed in this paper.

The subjects were studied in several hour sessions, where they had to perform various tasks and psychological exams. The final test was to do actual re-speaking of pre-recorded material. This was simultaneously recorded and recognized by a commercial, off-the-shelf ASR suite. The software was moderately adapted to the re-speaker during a several hour session, a few weeks before the test.

The materials included four different 5 minute segments in the speaker's native language and one in english (where the task was also translation). The recordings were additionally transcribed by a human, to convey exactly what the person said. The final dataset contains three sets of transcriptions:
1. the transcription of the original recorded material
2. the transcription of the re-speaker transcribed by a human
3. the output of the ASR recognizing the respeaker

# 5 Results Comparison

In our experiments we used 10 transcripts prepared using the protocol above. Each transcript was evaluated with all the metrics described in this paper as well as manually using the NER metric. Table 1 presents evaluation between human made text transcription and the original texts. Table 2 presents evaluation between ASR system and original texts.

| SPKR | BLEU | NIST | TER | METEOR | METEOR-PL | EBLEU | RIBES | NER | RED. |
|---|---|---|---|---|---|---|---|---|---|
| 1 | 56.20 | 6.90 | 29.10 | 79.62 | 67.07 | 62.32 | 85.59 | 92.38 | 10.15 |
| 2 | 56.58 | 6.42 | 30.96 | 78.38 | 67.44 | 58.82 | 86.13 | 94.86 | 17.77 |
| 3 | 71.56 | 7.86 | 18.27 | 88.28 | 76.19 | 79.58 | 92.48 | 94.71 | 12.01 |
| 4 | 76.64 | 8.27 | 13.03 | 90.34 | 79.29 | 87.38 | 93.07 | 93.1 | 3.72 |
| 5 | 34.95 | 5.32 | 44.50 | 61.86 | 47.74 | 37.06 | 71.60 | 91.41 | 17.03 |
| 6 | 61.73 | 7.53 | 20.47 | 83.10 | 72.55 | 69.11 | 92.43 | 92.33 | 4.89 |
| 7 | 61.74 | 6.93 | 28.26 | 78.26 | 69.78 | 63.99 | 78.32 | 95.3 | 10.29 |
| 8 | 33.52 | 4.28 | 46.02 | 63.06 | 47.61 | 36.75 | 77.55 | 93.95 | 26.81 |
| 9 | 68.97 | 7.46 | 22.50 | 83.15 | 76.56 | 71.83 | 88.78 | 94.73 | 4.05 |
| 10 | 70.02 | 7.80 | 18.78 | 86.12 | 78.71 | 75.16 | 88.15 | 95.23 | 6.41 |
| 11 | 47.07 | 5.56 | 33.84 | 76.10 | 62.02 | 47.86 | 85.05 | 93.61 | 22.77 |
| 12 | 53.49 | 6.65 | 30.63 | 77.93 | 65.27 | 55.90 | 86.20 | 94.09 | 14.33 |
| 13 | 75.71 | 7.95 | 16.07 | 89.72 | 83.13 | 77.74 | 91.62 | 94.78 | 9.31 |
| 14 | 66.46 | 7.60 | 18.44 | 84.34 | 76.20 | 66.53 | 90.76 | 94.82 | 6.09 |
| 15 | 25.77 | 1.85 | 54.65 | 58.62 | 40.82 | 31.08 | 68.90 | 85.26 | 32.83 |
| 16 | 88.82 | 8.66 | 5.75 | 96.15 | 90.12 | 95.20 | 95.76 | 95.96 | 2.71 |
| 17 | 63.26 | 7.25 | 25.72 | 81.95 | 73.16 | 65.83 | 88.99 | 94.77 | 10.15 |
| 18 | 60.69 | 7.18 | 26.23 | 79.41 | 70.86 | 66.77 | 87.56 | 95.15 | 5.75 |
| 19 | 59.13 | 7.2 | 25.04 | 80.00 | 70.32 | 62.77 | 89.17 | 95.78 | 4.74 |
| 20 | 86.24 | 8.43 | 7.11 | 94.60 | 90.07 | 92.88 | 95.39 | 95.58 | 1.52 |
| 21 | 20.61 | 2.08 | 65.14 | 47.56 | 33.31 | 27.42 | 55.63 | 91.62 | 36.89 |
| 22 | 64.40 | 7.43 | 22.84 | 82.69 | 72.82 | 66.93 | 89.44 | 93.46 | 10.15 |
| 23 | 27.30 | 2.69 | 52.62 | 58.96 | 43.19 | 35.78 | 68.06 | 90.07 | 31.81 |
| 24 | 82.43 | 8.33 | 10,15 | 92.40 | 86.04 | 85.61 | 94.63 | 97.18 | 12.02 |
| 25 | 82.22 | 8.44 | 9.48 | 93.75 | 87.00 | 91.59 | 95.12 | 97.01 | 3.89 |
| 26 | 76.17 | 8.25 | 13.54 | 91.33 | 81.54 | 82.35 | 91.95 | 95.83 | 8.63 |
| 27 | 35.01 | 4.39 | 49.41 | 62.64 | 47.81 | 46.64 | 72.79 | 87.56 | 25.21 |
| 28 | 29.50 | 3.53 | 53.13 | 60.73 | 42.98 | 40.87 | 74.25 | 85.76 | 28.09 |
| 29 | 70.04 | 7.78 | 17.26 | 87.22 | 80.63 | 73.58 | 91.56 | 93.98 | 8.63 |
| 30 | 56.75 | 6.89 | 26.06 | 79.60 | 70.52 | 58.43 | 90.78 | 95.79 | 11.68 |
| 31 | 63.18 | 6.90 | 26.57 | 83.47 | 71.80 | 67.80 | 84.51 | 94.29 | 17.77 |
| 32 | 31.74 | 5.14 | 43.15 | 63.58 | 49.37 | 33.07 | 83.05 | 94.77 | 19.12 |
| 33 | 89.09 | 8.54 | 5.41 | 95.69 | 91.62 | 92.62 | 95.94 | 97.41 | 0.34 |
| 34 | 81.04 | 8.42 | 8.29 | 93.60 | 88.51 | 89.68 | 95.18 | 93.76 | 5.75 |
| 35 | 73.72 | 8.11 | 15.40 | 88.62 | 78.32 | 83.17 | 92.48 | 93.89 | 4.40 |
| 36 | 69.73 | 7.90 | 15.06 | 87.90 | 81.65 | 78.10 | 93.91 | 93.97 | 7.45 |
| 37 | 57.00 | 7.24 | 26.40 | 81.28 | 68.54 | 65.83 | 86.63 | 92.74 | 8.46 |
| 38 | 35.26 | 3.68 | 46.70 | 66.04 | 47.84 | 40.73 | 74.61 | 89.15 | 27.07 |
| 39 | 46.76 | 4.92 | 39.59 | 72.95 | 57.66 | 54.28 | 72.77 | 88.07 | 25.21 |
| 40 | 16.79 | 1.74 | 65.48 | 44.61 | 29.04 | 19.62 | 58.77 | 86.68 | 32.66 |
| 41 | 19.68 | 0.63 | 61.42 | 56.13 | 39.15 | 41.59 | 49.67 | 84.36 | 52.12 |
| 42 | 39.19 | 5.04 | 41.62 | 68.38 | 50.85 | 42.41 | 79.79 | 91.23 | 23.35 |
| 43 | 67.13 | 7.61 | 19.12 | 86.53 | 75.63 | 68.60 | 93.51 | 95.01 | 11.84 |
| 44 | 49.85 | 6.26 | 33.84 | 75.31 | 64.34 | 59.44 | 79.28 | 92.23 | 11.84 |
| 45 | 37.38 | 4.2 | 43.49 | 68.66 | 54.30 | 44.51 | 75.36 | 93.11 | 27.58 |
| 46 | 79.11 | 8.25 | 12.01 | 91.48 | 85.72 | 81.66 | 94.84 | 95.36 | 6.09 |
| 47 | 40.73 | 4.93 | 41.96 | 68.61 | 54.34 | 42.96 | 83.09 | 89.97 | 21.15 |
| 48 | 29.03 | 2.65 | 51.27 | 61.24 | 47.21 | 39.53 | 67.21 | 86.07 | 31.98 |
| 49 | 68.75 | 7.78 | 18.78 | 86.24 | 77.18 | 72.40 | 90.65 | 94.95 | 5.41 |
| 50 | 75.24 | 7.97 | 16.07 | 88.88 | 81.51 | 81.27 | 90.88 | 95.75 | 7.45 |
| 51 | 78.71 | 8.24 | 11.51 | 91.33 | 83.99 | 86.08 | 94.77 | 98.69 | 4.23 |
| 52 | 37.60 | 4.31 | 44.84 | 66.32 | 51.59 | 42.22 | 78.73 | 89.37 | -6.6 |
| 53 | 73.20 | 8.07 | 14.38 | 88.78 | 79.55 | 77.39 | 93.93 | 93.73 | 2.2 |
| 54 | 67.43 | 7.67 | 20.30 | 85.64 | 75.90 | 70.28 | 87.57 | 94.91 | 12.07 |
| 55 | 70.06 | 7.90 | 18.44 | 87.29 | 76.67 | 78.93 | 90.79 | 93.49 | 7.11 |
| 56 | 71.88 | 7.83 | 17.77 | 88.24 | 78.07 | 77.51 | 89.21 | 97.74 | 8.46 |
| 57 | 80.05 | 8.3 | 11.00 | 91.81 | 85.72 | 83.94 | 95.03 | 96.12 | 2.88 |

**Table 1:** Evaluation of human made text transcription and original text (RED – reduction rate).

| SPKR | BLEU | NIST | TER | METEOR | METEOR-PL | EBLEU | RIBES | NER | RED. |
|---|---|---|---|---|---|---|---|---|---|
| 1 | 41.89 | 6.05 | 44.33 | 66.10 | 54.05 | 44.77 | 78.94 | 92.38 | 10.15 |
| 2 | 48.94 | 5.94 | 37.39 | 71.14 | 60.24 | 49.79 | 81.29 | 94.86 | 17.77 |
| 3 | 57.38 | 7.11 | 27.24 | 78.41 | 67.08 | 62.87 | 89.42 | 94.71 | 12.01 |
| 4 | 59.15 | 7.07 | 27.24 | 77.21 | 67.94 | 65.31 | 87.71 | 93.1 | 3.72 |
| 5 | 26.08 | 4.57 | 55.33 | 52.33 | 39.14 | 26.89 | 69.22 | 91.41 | 17.03 |
| 6 | 44.17 | 6.32 | 36.38 | 69.16 | 60.15 | 47.97 | 86.04 | 92.33 | 4.89 |
| 7 | 51.79 | 6.39 | 34.86 | 71.42 | 65.47 | 52.31 | 79.19 | 95.3 | 10.29 |
| 8 | 22.03 | 3.17 | 61.93 | 45.27 | 33.90 | 22.14 | 59.93 | 93.95 | 26.81 |
| 9 | 52.35 | 6.09 | 39.93 | 68.02 | 63.07 | 53.87 | 78.53 | 94.73 | 4.05 |
| 10 | 54.44 | 6.65 | 33.50 | 73.16 | 65.42 | 57.28 | 82.11 | 95.23 | 6.41 |
| 11 | 65.95 | 7.57 | 19.63 | 84.68 | 76.30 | 72.76 | 92.45 | 97.01 | 3.89 |
| 12 | 59.12 | 7.26 | 24.53 | 81.63 | 69.57 | 61.59 | 89.13 | 95.83 | 8.63 |
| 13 | 17.08 | 2.96 | 68.19 | 42.14 | 30.55 | 21.97 | 59.69 | 85.76 | 28.09 |
| 14 | 49.78 | 6.53 | 32.32 | 72.88 | 64.10 | 51.98 | 86.56 | 93.98 | 8.63 |
| 15 | 46.01 | 6.3 | 34.69 | 71.10 | 61.70 | 46.11 | 87.96 | 95.79 | 11.68 |
| 16 | 35.50 | 5.03 | 44.33 | 65.64 | 50.58 | 36.53 | 79.16 | 93.61 | 22.77 |
| 17 | 34.42 | 4.51 | 56.01 | 52.80 | 41.15 | 34.17 | 63.30 | 94.09 | 14.33 |
| 18 | 58.58 | 6.95 | 28.93 | 77.96 | 69.47 | 59.22 | 85.73 | 94.78 | 9.31 |
| 19 | 49.06 | 6.50 | 31.64 | 72.94 | 63.35 | 47.49 | 85.64 | 94.82 | 6.09 |
| 20 | 19.86 | 2.58 | 65.48 | 46.48 | 31.29 | 21.38 | 60.96 | 85.26 | 32.83 |

**Table 2**: Evaluation between ASR and original text

To find the most reliable metrics in place of NER, a backward linear regression has been used [18]. This regression has been selected for the following reasons:

1  The data are linear data (correlation analysis present linear relation)
2  The data are ratio level data, thus good for linear regression
3  This regression analysis provides more than one regression results, thus the best variables can be found
4  Reliable variables are extracted by excluding irrelevant variables from the 1st to the last stage (give the most reliable variables).

For this analysis, the Standardized Coefficients would be seen as strength of relationship that extracted from the Unstandardized Coefficients [19]. The sig (p-value) would be judged with the alpha (0.05) to investigate the most significant variables that explain the NER metrics. The Adjusted R-square ($R^2$) [20] would provide the idea, how much the variables are explaining the NER variances.

Table 3 represents the regression summary for NER and other metrics. Here, at first, the model has all the metrics and except EBLEU none of them are significant predictors of NER. Additionally, among all the metrics, TER is the most insignificant metric, thus it has been removed for the second model. In the second model only EBLEU (p = 0.007) and NIST (p = 0.019) were significant as they were below the alpha (0.05). Here in the 2nd model the most insignificant metrics is RIBES (p = 0.538), thus it was removed in the 3rd model at the next stage. In the 3rd model, again EBLEU (p = 0.006) and NIST (p = 0.018) is significant and their significance value increased so does the Beta for EBLEU increased, the stronger they become in predicting NER. In this model, METEOR_pl is the most insignificant metric, thus it has been removed for the next model.

In the 4th model, EBLEU, BLEU and NIST are significant and the level of significance for EBLEU and BLEU has increased as well as their beta values, nonetheless the level of significance for NIST remain same. However, at this stage of the model, METEOR become the most insignificant metric, thus it has been removed for the final (5th model). Along, side compared to 2nd and, 3rd model, in 4th model it has been seen that the BLEU is becoming more significant as the p-value (p = 0.005) is less than the alpha (0.05). Finally, the last stage (5th model) the remaining metrics are significant, as all have p-values less than alpha. Thus, no next step of models has been executed. Here, the fist model explaining 75.9% of the variance for NER, however, the second model explaining 76.3% of the variance of the NER, 3rd model explaining

76.6% of the variance of the NER, 4th model explaining 76.0% of the variance of the NER and final model explaining 76.1% of the variance of the NER. All these are statistically accepted R-square values.

| Model | | Unstandardized Coefficients | | Standardized Coefficients | t | Sig. | Adjusted R-square |
|---|---|---|---|---|---|---|---|
| | | B | Std. Error | Beta | | | |
| 1st model | (Constant) | 100.068 | 19.933 | | 5.020 | .000 | 0.759 |
| | BLEU | .176 | .122 | 1.060 | 1.439 | .157 | |
| | NIST | 1.162 | .530 | .738 | 2.192 | .033 | |
| | TER | -.072 | .186 | -.353 | -.386 | .701 | |
| | METEOR | -.205 | .168 | -.797 | -1.216 | .230 | |
| | EBLEU | -.218 | .078 | -1.293 | -2.806 | .007 | |
| | RIBES | -.067 | .093 | -.219 | -.723 | .473 | |
| | METEOR_pl | .194 | .160 | .954 | 1.216 | .230 | |
| 2nd model | (Constant) | 92.622 | 4.895 | | 18.920 | .000 | 0.763 |
| | BLEU | .189 | .117 | 1.136 | 1.614 | .113 | |
| | NIST | 1.220 | .504 | .775 | 2.420 | .019 | |
| | METEOR | -.190 | .163 | -.741 | -1.170 | .248 | |
| | EBLEU | -.215 | .077 | -1.270 | -2.803 | .007 | |
| | RIBES | -.051 | .082 | -.167 | -.620 | .538 | |
| | METEOR_pl | .217 | .147 | 1.067 | 1.479 | .145 | |
| 3rd model | (Constant) | 91.541 | 4.547 | | 20.133 | .000 | 0.766 |
| | BLEU | .184 | .116 | 1.107 | 1.586 | .119 | |
| | NIST | 1.068 | .438 | .678 | 2.440 | .018 | |
| | METEOR | -.237 | .143 | -.923 | -1.655 | .104 | |
| | EBLEU | -.193 | .068 | -1.145 | -2.841 | .006 | |
| | METEOR_pl | .222 | .146 | 1.093 | 1.527 | .133 | |
| 4th model | (Constant) | 90.471 | 4.550 | | 19.885 | .000 | 0.760 |
| | BLEU | .285 | .096 | 1.716 | 2.955 | .005 | |
| | NIST | 1.083 | .443 | .687 | 2.444 | .018 | |
| | METEOR | -.099 | .112 | -.385 | -.878 | .384 | |
| | EBLEU | -.204 | .069 | -1.210 | -2.982 | .004 | |
| 5th model | (Constant) | 86.556 | .913 | | 94.814 | .000 | 0.761 |
| | BLEU | .254 | .090 | 1.531 | 2.835 | .006 | |
| | NIST | .924 | .404 | .587 | 2.289 | .026 | |
| | EBLEU | -.221 | .066 | -1.310 | -3.370 | .001 | |

a. Dependent Variable: NER

**Table 3:** Regression result summary for NER and the six metrics

# 6 Conclusion

From the regression it has been found that BLEU is the most significant predictor of NER, after BLEU, NIST is significant and finally, EBLEU is also a significant metric that can predict the NER better and thus these can be alternative to the NER metric. The regression equation that can compute the value of NER based on these three statistical metrics as:

$$NER = 86.55 + 0.254 * BLEU + 0.924 * NIST - 0.221 * EBLEU$$

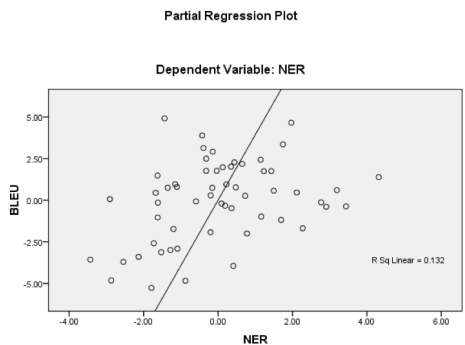

**Figure 1:** Partial regression plot for NER and BLEU

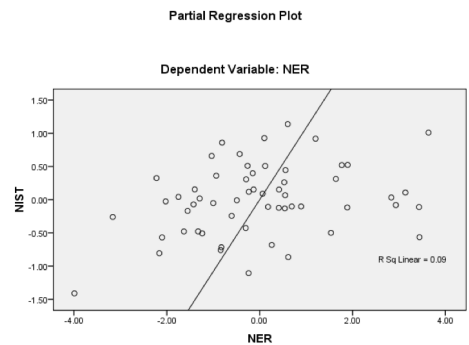

**Figure 2:** Partial regression plot for NER and NIST

Moving forward, regression plots from the above regression for the significant metrics are presented below. These plots show the relation between dependent metric with each significant metric. The closer the dots in the plot to the regression line the better the R-square value and the better the relationship.

It is worth noting that the METEOR metric should be fine-tuned to work properly. This would however require more data and more research into this is planned in the future. Finally, the results presented in this paper are derived from a very small data sample and may not be very representative of the task in general. The process of acquiring more data is still ongoing so these experiments are going to be repeated once more data becomes available.

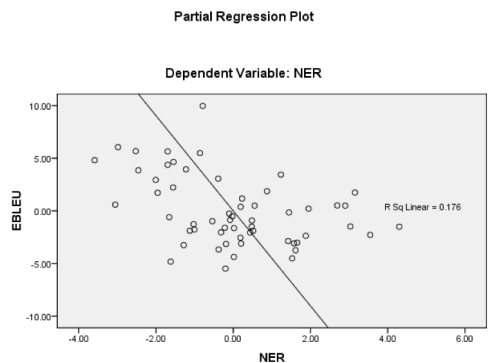

**Figure 3:** Partial regression plot for NER and EBLEU

## Acknowledgements

We would like to thank dr Agnieszka Szarkowska and Łukasz Dutka for sharing the data that allowed us to perform these experiments.

Some of the research used in this paper was funded by the CLARIN-PL project.